\documentclass[sigconf,10pt,nonacm]{acmart}
\usepackage{multirow}
\usepackage{xspace}
\usepackage{listings}
\usepackage{subcaption}
\usepackage{tikz}
\usepackage{paralist}
\usepackage{cleveref}
\usepackage{tcolorbox}
\usepackage{xcolor}
\usepackage{amsthm}
\usepackage{thmtools}

\newcommand{\fulg}{FuLG\xspace}

\definecolor{gray}{HTML}{ededed}
\colorlet{shadecolor}{gray}
\tcbset{
	colback=gray,
	boxrule=0.5pt,
	boxsep=1pt,
	left=3pt,
	right=3pt,
	top=3pt,
	bottom=3pt,
	sharp corners,
	colframe=black,
}

\newcommand{\headline}[1]{\vspace{+0.2cm}\noindent\textbf{#1}}

\declaretheoremstyle[
    spaceabove=6pt, 
    spacebelow=6pt,
    headfont=\normalfont\bfseries,
    notefont=\mdseries, 
    notebraces={(}{)},
    bodyfont=\normalfont,
    postheadspace=1em,
    headpunct={:},
    headformat=\NAME\NOTE,
]{defstyle}

\title{\fulg: 150B Romanian Corpus for Language Model Pretraining}

\author{Vlad-Andrei Bădoiu}
\affiliation{
    \institution{University Politehnica of Bucharest}
    \city{Bucharest}
    \country{Romania}
}

\author{Mihai-Valentin Dumitru}
\affiliation{
    \institution{University Politehnica of Bucharest}
    \city{Bucharest}
    \country{Romania}
}

\author{Alexandru M. Gherghescu}
\affiliation{
    \institution{University Politehnica of Bucharest}
    \city{Bucharest}
    \country{Romania}
}

\author{Alexandru Agache}
\affiliation{
    \institution{University Politehnica of Bucharest}
    \city{Bucharest}
    \country{Romania}
}

\author{Costin Raiciu}
\affiliation{
  \institution{University Politehnica of Bucharest}
   \institution{Broadcom Inc.}
    \city{Bucharest}
    \country{Romania}
}

\begin{abstract}

Research in the field of language models is rapidly evolving, with many open
models being released to the public. Openly available pretraining corpora
usually focus on only a handful of languages, with many others  either
missing completely or extremely underrepresented. In this report, we
introduce \fulg~\footnote{\url{https://hf.co/datasets/faur-ai/fulg}}, a hundred-fifty-billion-token Romanian corpus extracted from
CommonCrawl. We present our methodology for filtering \fulg and compare it via
ablation studies against existing Romanian corpora. 

\end{abstract}

\begin{document}

\maketitle

\section{Introduction}

In a world where Large Language Models (LLMs) have shown great potential for
natural language tasks, the corpora used for pretraining play a central role in
their overall capabilities. However, most state-of-the-art models are not
accompanied by their training datasets and authors only sparsely discuss their
composition.

In response to this, researchers and practitioners have produced numerous data
corpora for widely spoken languages on the Internet. These are often derived
from CommonCrawl~\footnote{\url{https://commoncrawl.org}}, a public repository of crawled web pages, which has indexed
more than 250 billion pages. Other datasets go further by including curated
data such as books, social media discussions, and research papers.
Trillion-token datasets such as Dolma~\cite{dolma}, FineWeb~\cite{fineweb}, or
RedPajama~\cite{redpajama} have enabled the development of billion-parameter
models with loss-optimal training, but they cover only a small fraction of
languages.

This limitation is evident in the language understanding capabilities of open
models, as they are often trained on publicly available data. For instance,
OLMo models~\cite{olmo} support only six languages, while other initiatives such as
LLM360 K2~\cite{llm360} are English-centric. Even models developed by tech giants, such as
Llama~\cite{llama} or Gemma~\cite{gemma}, have limited output capabilities in less commonly spoken
languages. Only recently have we seen some developments in this area, with
Llama 3 supporting 30 languages, albeit with lower performance levels compared
to English and limited information about the training dataset.

To improve the standing of the Romanian language in future models, we delve
into filtering Romanian content from CommonCrawl, documenting the process and
releasing an open dataset. We built a pipeline that employs deduplication
techniques, common signal filters, and FastText for language detection. We
document the steps taken to obtain the final text corpus. As a result, we
release \fulg, an openly available corpus that is three times larger than
existing Romanian corpora. 

In this work we make two main contributions. We release \fulg, a 156B-token
(589GB tokenized), or 220B tokens with the Llama 3 tokenizer, corpus for LLM
pretraining and fine-tuning in the Romanian language. Second, we document the
process of filtering data from CommonCrawl, which may be employed for other
underrepresented languages.

\section{Related Work}

A crucial factor in model performance is the size, quality and diversity of
the pretraining corpus. These datasets typically comprise content from various
sources, primarily web pages, due to their sheer number and availability, followed
by social media discussions, academic papers, books, code repositories, and
encyclopedias, covering a wide range of topics. The main sources of data are either
private crawls of the Internet or the publicly available CommonCrawl repository.

While information about closed models' training data is scarce, even open
models like Llama or Gemma lack transparency regarding their training datasets,
as it is a central part of the competitive advantage. We are thus in a peculiar
scenario, where the best open-source models aren't truly open-source when it
comes to pretraining data, which deeply affects any smaller initiative.

On the other hand, open
datasets are essential components for truly open models and play a vital role
in democratizing future LLMs. Currently, numerous open
datasets exist, many derived from CommonCrawl. 
Notable examples include: Dolma~\cite{dolma} (3T tokens), C4~\cite{C4} (175B tokens), The
Pile~\cite{pile} (387B tokens), ROOTS~\cite{roots} (400B tokens),
RefinedWeb~\cite{refinedweb} (600B tokens), RedPajama v2~\cite{redpajama} (30T
tokens), FineWeb~\cite{fineweb} (15T tokens), Zyda~\cite{zyda} (1.3T tokens),
LLM360 Amber~\cite{llm360} (1.2T tokens), MAP-Neo~\cite{map-neo} (4.5T tokens), OSCAR~\cite{oscar}.
While these datasets provide sufficient quality and size for English-based
models, they often lack adequate representation of less commonly spoken
languages. For example, after training a Romanian GPT-NeoX tokenizer on the Romanian part of OSCAR,
we obtain a number of only 10B tokens on OSCAR, and 41B tokens on
mC4~\cite{mC4} (slightly higher number of tokens if using an
English tokenizer, similar to Llama or OLMo, due to worse compression ratio). However, considering the scaling
laws of dataset size relative to model size, these quantities are insufficient
for optimal LLM training. Beyond these, several smaller datasets exist for
various Natural Language Processing (NLP) tasks in Romanian~\cite{marcell,
corola, rombac, midrigan2020resources, eltec, manolescu2021roff} but they
generally contain less than 500M tokens, which is far from adequate for LLM
development.

\section{Data Acquisition and Filtering}

\fulg is derived from CommonCrawl, a vast collection of web page crawls dating
back to 2007. CommonCrawl releases snapshots of portions of the Internet
several times annually. These snapshots exhibit low similarity rates between
releases, enabling us to consistently extract new data from each snapshot. Our work
encompasses snapshots from 2013 to May 2024.

We developed a pipeline based on existing software, which we elaborate on
shortly. Given the petabytes of data requiring filtration, we utilized a
distributed environment with multiple nodes for data acquisition. For
deduplication and quality filtering, we employed a single large-memory node.

\paragraph{Data Acquisition} To process CommonCrawl snapshots, we leveraged the
CCNet pipeline~\cite{ccnet}. CCNet facilitates distributed processing of
snapshots, including downloading in WET format, language identification via the
FastText algorithm~\cite{fasttext}, and deduplication of common paragraphs. We
encountered two primary challenges with CCNet:

\begin{enumerate}

\item Due to the fact that CCNet development has stalled, we quickly hit
roadblocks related to package versions and environment-specific problems
with our SLURM setup, which needed modifications to the source code.
\item Since we ran the pipeline on a cluster shared with other users, we did not
have all the hardware for ourselves. Strict limits imposed by the SLURM system
for fair sharing meant we needed to adapt the source code, which was not built
with this in mind. We faced limitations with the maximum size of job arrays
supported, often capped at 100. To address this, we
extended CCNet to support job batching. Initially, CCNet would submit a
job array equal in size to the number of shards; our modifications
enabled it to break the job array into multiple smaller submissions to the SLURM
scheduler.

\end{enumerate}

We kept only documents with a language score for Romanian exceeding 0.5,
meaning that there are documents that may include other languages alongside
Romanian.

\paragraph{Deduplication}
For filtering and deduplication, we employed code from
RedPajama~\cite{redpajama}. Exact deduplication reduced the dataset size by
37\%, while fuzzy deduplication with a 0.8 threshold further reduced it by
50\%. We updated the deduplication pipeline to not only identify duplicates but
also remove them from the dataset.

\paragraph{Content Filtering}
Next, we introduced a content filtering step to our pipeline. Using a
regex-based approach, we filtered HTML text extraction artifacts, such as
navbar text, from documents. To filter potentially controversial content, we utilized a dictionary,
removing documents containing specific words in either the content or URL. For
Personally Identifiable Information (PII), we replaced phone numbers, email
addresses, and links with special tokens.

\paragraph{Quality Filtering}
As with deduplication, we utilized existing code from RedPajama to compute a
set of quality signals. We then extended the code to filter documents based on
the quality signals introduced by Gopher~\cite{gopher}, reducing the dataset
size by 50\% to 156B tokens using the OLMo tokenizer trained on Romanian.
(589GB). With a different tokenizer such as Llama 3, which was trained on many
other languages, we obtain around 220B tokens; we expect much worse compression
for English-only tokenizers, such as Llama 2. The thresholds and rules we
employed for filtering are as follows:

\begin{itemize}
\item Fraction of characters in the most common n-grams exceeded: 0.2 for
2-grams, 0.18 for 3-grams, 0.16 for 4-grams, 0.15 for 5-grams, 0.14 for
6-grams, 0.13 for 7-grams, 0.12 for 8-grams, 0.11 for 9-grams, 0.10 for
10-grams
\item Fewer than 50 words or more than 100,000 words
\item Median word length less than 3 or greater than 10 characters
\item More than 90\% of lines start with bullet points
\item More than 30\% of lines end with ellipses
\item Less than 30\% of lines end with punctuation
\end{itemize}

The data acquisition process spanned several months, while deduplication and
filtering were completed in less than 18 hours on a single large-memory node.


\section{F\MakeLowercase{u}LG in action}


To evaluate \fulg we conducted ablation studies using a 1B decoder-only model based on
OLMo~\cite{olmo}, with a sequence length of 2048 and the OLMo tokenizer trained
on the OSCAR-Ro corpus. We used a global batch size of 256. We made several changes to
the hyperparameters to adjust to the 1B model size: weight decay of 0.001 and
a max learning rate of 4e-4 annealed by a cosine schedule, with an initial warmup
of 1000 steps.

We trained three identical models on different pretraining datasets: Oscar, mC4
and \fulg . We leveraged the Hugging Face transformers~\cite{transformers}
library for training. We trained to completion on each dataset (290k steps for
\fulg, 75k for mC4 and 20k for OSCAR). Since \fulg was much bigger than both
OSCAR and mC4 (about 4 times bigger than mC4), we also included an earlier
checkpoint from \fulg at 70k steps. We do note that it slightly underperforms,
but we believe this naturally happens because of the cosine decay learning rate schedule.


\begin{table}[h]
    \centering
    \begin{tabular}{|c|c|}
        \hline
        Dataset & Perplexity\\
        \hline
        \fulg (290k) & 16.06 \\
        \fulg (70k) & 21 \\
        mC4 & 15.38 \\
        OSCAR & 23.57\\
        \hline
    \end{tabular}
    \caption{Perplexities for datasets.}
    \label{tab:perplexities}
\end{table}

\headline{Perplexity.} A commonly employed method to assess the quality of a
dataset is to fix a model, train it with different datasets, and measure
perplexity against a curated evaluation set. While perplexity is not a
definitive measure of dataset quality~\cite{dolma}, we use it as a sanity
check, with values similar to existing datasets confirming our approach. We
constructed a perplexity dataset covering multiple domains, sourced from
Wikipedia, news articles, textbooks, research papers, and books, totaling 74M
tokens. Although we did not specifically perform a separate decontamination
step, as this is only a preliminary evaluation, we do note that the collection
methods of the evaluation dataset were manual, with care in picking high-quality
sources. We do plan to apply a decontamination step in further experiments, to
make sure our results are fair. Before calculating perplexity, the evaluation
dataset was cleaned using the \textit{clean-text} Python package
\footnote{\url{https://github.com/jfilter/clean-text}}. The results are shown
in Table~\ref{tab:perplexities}. Both ablations of \fulg show perplexity values
similar to those of existing datasets.

\begin{table}[h]
    \centering
    \begin{tabular}{|c|c|c|c|}
        \hline
        Dataset & Grammar & Creativity & Complexity \\
        \hline
        \fulg (290k) & 8.5 & 7.1 & 4.6 \\
        \fulg (70k) & 7.25 & 5.875 & 4 \\
        mC4 & 7.5 & 5.2 & 3.1 \\
        OSCAR & 6.3 & 4.2 & 2.8 \\
        \hline
    \end{tabular}
    \caption{Results for the story generation task.}
    \label{tab:dataset_characteristics}
\end{table}

\headline{Story generation} We further qualitatively evaluated the trained
models. Inspired by the TinyStories~\cite{tinystories} work, we used the models
to generate stories from given prompts and asked GPT-4 to rate the creativity,
grammar, and overall complexity of each story. We only considered responses
that were coherent stories, discarding anomalous outputs. Both OSCAR and \fulg
(70k) generated significantly more anomalies than the other ablations. In
Table~\ref{tab:dataset_characteristics}, we present our findings, which suggest
hat \fulg may enable better performance for specific tasks.

\section{Discussion and Future Work}

The availability of high-quality datasets for less commonly spoken languages is
crucial for the democratization of LLMs. While proprietary models often
demonstrate proficiency across a wide range of languages, open models
frequently underperform in this aspect. By developing large, high-quality
open datasets for diverse languages, we can foster the creation of superior
open-source models. These improved models could potentially be utilized by
governments worldwide to meet their digitalization needs.

\fulg represents
an initial effort to improve dataset size and quality for the Romanian
language. We're already underway to implementing a few other improvements which
will further increase dataset quality:


\begin{itemize}

\item Data processing: We currently work with Common Crawl snapshots in WET format.
However, as previous research [citation] indicates, employing a more
sophisticated HTML parser could significantly improve both the quality of
extracted text and the overall volume of usable data.

\item Language-specific optimization: Romanian language particularities could be
leveraged to adapt and refine quality filters. Currently, we use thresholds
designed for the English language, but developing language-specific criteria
could yield better results.

\item Novel quality filters: Identifying and implementing new quality filters
tailored to the Romanian language could further enhance the dataset's overall
quality and relevance.

\end{itemize}

\section*{Acknowledgements}
We acknowledge the EuroHPC Joint Undertaking for awarding this project access
to the EuroHPC supercomputer LUMI, hosted by CSC (Finland) and the LUMI
consortium through a EuroHPC Regular Access call. Some of the experiments were
performed on the Luxembourg national supercomputer MeluXina. The authors
gratefully acknowledge the LuxProvide teams for their expert support. We
acknowledge the computational resources provided by the PRACE award granting
access to Discoverer in SofiaTech, Bulgaria. The authors gratefully acknowledge
the HPC RIVR consortium and EuroHPC JU for funding this research by providing
computing resources of the HPC system Vega at the Institute of Information
Science. We acknowledge VSB – Technical
University of Ostrava, IT4Innovations National Supercomputing Center, Czech
Republic, for awarding this project access to the LUMI supercomputer, owned by
the EuroHPC Joint Undertaking, hosted by CSC (Finland) and the LUMI consortium
through the Ministry of Education, Youth and Sports of the Czech Republic
through the e-INFRA CZ (grant ID: 90254). This work was funded by a VMWare gift.

\bibliographystyle{ACM-Reference-Format}
\bibliography{ref}

\end{document}